\documentclass[11pt]{article}
\usepackage{acl}
\usepackage{times}
\usepackage{latexsym}
\usepackage{amsmath}
\usepackage{amssymb}
\usepackage{booktabs}
\usepackage{graphicx}
\usepackage{multirow}
\usepackage{microtype}
\usepackage{inconsolata}
\usepackage{bbm}

\title{STEP: Soft Trajectory Estimation of Prediction Correctness \\ via Per-Step Probability Scoring}
\author{%
  Chenyang Yu\textsuperscript{1}\thanks{\ Work done during an internship at the RAMS Lab, Huawei Heisenberg Research Center.} \quad
  Zifan Zeng\textsuperscript{1,2} \quad
  Qunli Zhang\textsuperscript{2} \quad
  Feng Liu\textsuperscript{2} \\[4pt]
  \normalsize\normalfont
  \textsuperscript{1}Technical University of Munich \\
  \normalsize\normalfont
  \textsuperscript{2}RAMS Lab, Huawei Heisenberg Research Center \\[4pt]
  \small
  \texttt{chenyang036.yu@tum.de}, \quad
  \texttt{zifan.zeng@huawei.com}, \quad
  \texttt{zhangqunli1@huawei.com}, \quad
  \texttt{feng.liu1@huawei.com}
  }

\begin{document}
\maketitle

\begin{abstract}
Estimating the correctness of large language model (LLM) predictions without 
access to ground truth is critical for reliable deployment in high-stakes settings. 
Existing approaches, such as entropy and perplexity, operate solely on final output 
statistics, discarding the rich information encoded in intermediate reasoning steps. 
To address this limitation, we propose \textit{\textbf{STEP} (\textbf{S}oft 
\textbf{T}rajectory \textbf{E}stimation of \textbf{P}rediction correctness)}, a 
training-free framework that scores each reasoning step via lightweight 
forward passes to obtain a soft probability distribution $q_i$ over candidate 
answers, from which we derive a set of \textit{soft trajectory features} capturing 
per-step confidence and its progression toward the final answer. Extensive experiments on various benchmarks demonstrate consistent and substantial improvements, particularly in AUROC, over output-based baselines and hard-indicator trajectory methods. 
Notably, \textit{Temporally-Weighted Consistency}, which assigns higher 
importance to later reasoning steps where answer distributions have stabilized, 
contributes the largest gains, highlighting the importance of temporal structure in 
reasoning trajectories. These results further validate the advantage of preserving 
continuous probability signals over binary discretization.
\end{abstract}

\section{Introduction}

Large language models (LLMs) have demonstrated remarkable capabilities across a 
wide range of reasoning tasks, from mathematical problem solving to multi-step 
commonsense inference. Despite their impressive performance, these models remain 
prone to generating incorrect answers with high apparent confidence, making 
correctness estimation at inference time a critical requirement for reliable 
deployment. It is essential, particularly in high-stakes settings such as medical 
question answering, legal reasoning, and automated decision support, where acting 
on a wrong prediction can carry serious consequences.

Existing methods typically rely on final output statistics derived from the model's
token-level distributions: entropy over the output \citep{malinin2021uncertainty, 
kuhn2023semantic}, token-level perplexity \citep{guo2017calibration}, or the 
probability assigned to the most likely answer \citep{hendrycks2017baseline}. 
Although these signals are simply accessible from the model's logits, they operate 
at the level of individual tokens and capture only local fluency and output sharpness 
\citep{guo2017calibration}. They do not reflect whether the model has engaged in 
coherent, logically grounded reasoning at a semantic level \citep{kuhn2023semantic}. 
A model can assign high probability to a wrong answer simply because that answer is 
superficially plausible \citep{kadavath2022language}; conversely, a correct answer 
may receive modest token-level probability when the model remains genuinely uncertain 
yet ultimately reaches the right conclusion. Without examining the reasoning process 
itself, such methods lack a principled connection to semantic correctness.

Another approach to correctness estimation is to dive into the latent space of the 
model. A line of work has shown that the internal representations of LLMs carry 
signals correlated with factual correctness, and lightweight probes trained on hidden 
states can reliably predict output correctness \citep{azaria2023internal, 
kossen2024sep}. However, even lightweight probes require labeled training data and 
are sensitive to distribution shift, limiting their applicability when the task 
domain or model changes. Our approach requires no such training, and is thus 
applicable out of the box across tasks and models.

Another line of work exploits the internal geometry of reasoning trajectories. 
\citet{wang2024coe} and \citet{bi2025cotkinetics} derive correctness signals from 
the trajectory of hidden states across layers, showing that correct and incorrect 
responses exhibit measurably different geometric patterns in embedding space. A 
related but distinct approach is taken by \citet{zhang2025covo}, who operate in 
the output probability space rather than the latent space: correct trajectories 
exhibit high consistency, where intermediate steps steadily converge toward the 
final answer, while incorrect trajectories deviate erratically before settling on 
a wrong conclusion. However, this framework has two limitations that prevent direct 
application to inference-time correctness estimation. First, it requires sampling 
$N$ independent trajectories per prompt, which scales inference cost proportionally. 
Second, correctness is measured via hard binary indicators that discard the full 
spectrum of confidence information, treating near-chance and near-certain 
probabilities identically.

In this work, we propose a training-free framework for correctness
estimation of LLM output in closed-set tasks, which represent a broad and
practically important class of real-world applications: from standardized
tests and professional certification exams to clinical decision support and
automated document classification, many high-stakes settings inherently present
a finite, predefined set of options. Inspired by the trajectory-based insight 
of \citet{zhang2025covo}, we address both limitations by re-scoring the 
already-generated reasoning trajectory: truncating it at each intermediate step 
$i$ and performing a lightweight forward pass to obtain a soft probability 
distribution $q_i$ over all candidate answers, requiring no additional generation 
and no resampling.

This per-step distribution is a strictly richer signal than a binary
correct/incorrect label: it captures not just which answer the model favors at
each step, but how confidently and consistently it does so across the full
trajectory. From these distributions, we derive a family of \textbf{soft 
trajectory features} reflecting distinct aspects of reasoning dynamics 
(Section~\ref{sec:method}).

We conduct extensive experiments across 4 datasets and 2 models, evaluating our 
features against output-based methods, training-free trajectory-based methods, 
and latent-trajectory approaches. Our soft features achieve consistent and substantial improvements, particularly in AUROC, across models and datasets. 
Our contributions are as follows:
\begin{enumerate}
  \item We propose \textbf{STEP}, a single-trajectory, training-free framework for correctness
    estimation in closed-set tasks, grounded in per-step soft probability
    distributions over the candidate answer set.
\item In STEP, we introduce three soft trajectory metrics that preserve continuous 
    confidence signals discarded by prior binary-indicator methods.
\item We provide systematic empirical evidence across four benchmarks and two
    open-source models that soft trajectory features achieve strong and consistent performance relative to existing baselines, and find that later reasoning steps carry
    disproportionately more predictive signal.
\end{enumerate}

\begin{figure*}[t]
    \centering
    \includegraphics[width=1\linewidth]{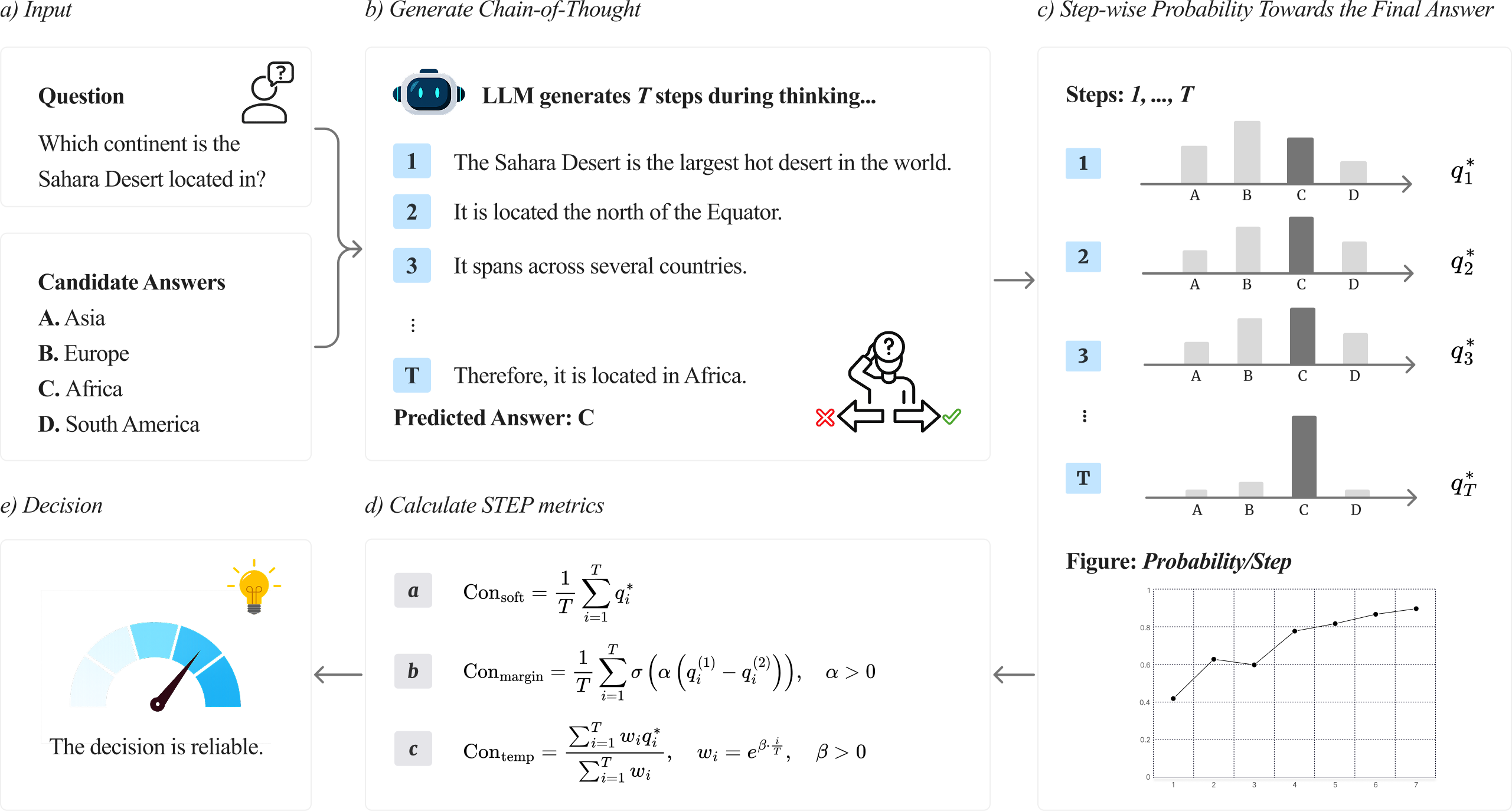}
    \caption{Overview of the STEP framework. (a) Given a question and a predefined 
    candidate answer set, (b) the LLM generates a $T$-step chain-of-thought followed 
    by a predicted answer. (c) For each prefix $c_i$, a lightweight forward pass 
    yields a soft probability distribution over candidate answers, from which $q_i^*$ 
    is extracted. The resulting trajectory $\{q_1^*, \ldots, q_T^*\}$ reflects how 
    confidently the model commits to its predicted answer across reasoning steps. 
    (d) Three soft trajectory features are computed from this trajectory. 
    (e) The resulting correctness score is used to assess the reliability of the 
    model's prediction.}
    \label{fig:pipeline}
\end{figure*}

\section{Related Work}

\paragraph{Output-statistics-based uncertainty estimation.}
A foundational line of work estimates output reliability from token-level
statistics derived from a single forward pass. Predictive entropy over the
output distribution~\citep{malinin2021uncertainty}, length-normalized
log-probability, and the maximum softmax probability~\citep{hendrycks2017baseline}
are canonical examples. These signals are cheap to compute but operate at the
level of individual tokens: they capture local fluency and output sharpness
rather than semantic correctness. A model can assign high token-level
probability to a wrong answer that is superficially fluent, and low probability
to a correct but uncertain answer~\citep{guo2017calibration}.

\paragraph{Semantic and sampling-based methods.}
To capture meaning-level uncertainty, \citet{kuhn2023semantic} introduce
\emph{semantic entropy}, which clusters multiple sampled responses by semantic
equivalence and computes entropy over the resulting distribution, substantially
improving over token-level baselines. \citet{farquhar2024detecting} extend this
to a practical hallucination detection method, demonstrating its effectiveness
at scale. However, both approaches require sampling $N$ responses per
prompt, which is a significant inference overhead that limits deployment in
latency-sensitive settings.

\paragraph{Probing-based methods.}
A complementary line of work shows that LLM hidden states encode signals
correlated with truthfulness and correctness. \citet{burns2023discovering} and
\citet{marks2024geometry} show that truthfulness is linearly represented in
activation space and can be recovered without supervision, by finding directions
that separate true from false statements. While these unsupervised methods
require no correctness labels, they still require fitting to the target
distribution and have been shown to generalize poorly to structurally different
inputs. Supervised probes face the same limitation more acutely:
\citet{azaria2023internal} train a classifier on hidden layer activations to
predict statement truthfulness, and \citet{kossen2024sep} train linear probes
to approximate semantic entropy from a single forward pass. In both the
supervised and unsupervised settings, the reliance on distribution-specific
fitting limits applicability when the task domain or model changes, which is a
limitation our training-free approach avoids entirely.

\paragraph{Trajectory-based correctness estimation.}
Several works exploit the structure of the reasoning process rather than the
final output alone. \citet{wang2024coe} show that the trajectory of hidden
states across layers exhibits measurable differences between correct and
incorrect responses: correct trajectories tend to travel greater distances in
embedding space and maintain more consistent direction, and their CoE-R and
CoE-C scores aggregate these geometric signals as training-free correctness
estimates. \citet{bi2025cotkinetics} model the dynamics of hidden state
transitions across reasoning steps from a theoretical perspective, deriving
a scalar score that reflects the soundness of the reasoning trajectory.
These methods operate entirely in the latent space and require white-box
access to specific layer activations.

A related but distinct approach is taken by \citet{zhang2025covo}, who operate
in the output probability space rather than the latent space. In their 
\emph{CoVo} framework, correct trajectories exhibit high \emph{consistency} 
(intermediate steps consistently favor the final answer) and low
\emph{volatility} (the last deviating step occurs early in the chain). Our 
method builds on this insight but addresses two key limitations: \textit{CoVo} requires 
sampling $N$ independent trajectories per prompt, and reduces per-step 
confidence to hard binary indicators, discarding fine-grained confidence 
information. We exploit the closed-set property of the task to operate on a 
single trajectory, and replace binary indicators with soft probability 
distributions that preserve the full confidence signal.

\paragraph{Chain-of-thought reasoning and faithfulness.}
\citet{wei2022chain} demonstrate that prompting LLMs to generate intermediate
reasoning steps substantially improves accuracy on multi-step tasks. A key
assumption underlying trajectory-based correctness estimation, including our
own, is that these intermediate steps are meaningful reflections of the model's
reasoning process. Our method exploits this structure by truncating the
chain-of-thought at each step and probing the model's implicit answer
preference, treating the reasoning trajectory as a window into the model's
evolving commitment to candidate answers. This is complementary to process
reward models~\citep{lightman2023lets}, which also exploit step-level
structure but require annotated training data; our approach is training-free
and derives correctness signals directly from the model's output distribution
at each step.
\section{Method}
\label{sec:method}

\subsection{Problem Formulation}

We focus on \textbf{closed-set reasoning tasks}, in which a model is given an 
input question $x$ and a predefined set of $M$ candidate answers 
$\mathcal{A} = \{a_1, a_2, \ldots, a_M\}$, and generates a chain-of-thought (CoT)
followed by a final answer $\hat{a} \in \mathcal{A}$.

Given the question $x$, candidate set $\mathcal{A}$, and the generated 
chain-of-thought $\mathcal{C}$, our goal is to compute a scalar correctness score 
$\phi(x, \mathcal{C}) \in \mathbb{R}$ that correlates with whether $\hat{a}$ matches 
the ground-truth answer, without access to labels. We evaluate methods using AUROC,  AUPR-C and AUPR-E (see Section~\ref{sec:setup} and Appendix~\ref{app:metrics} for 
formal definitions).

\subsection{Per-Step Answer Distribution}

\paragraph{Step Segmentation.} 
The model is prompted to generate reasoning in a structured format, with each 
logical step on a separate line. The resulting chain-of-thought $\mathcal{C}$ is 
segmented into $T$ discrete steps $\{s_1, s_2, \ldots, s_T\}$. The prefix context 
at step $i$ is defined as
\[
c_i = [x, s_1, s_2, \ldots, s_i], \quad i = 1, \ldots, T.
\]

Each prefix $c_i$ represents the model's reasoning state after $i$ steps: it 
captures everything the model has considered up to that point, without revealing 
how the reasoning will continue. By probing the model at each such prefix, we 
obtain a snapshot of its implicit answer preference at every stage of the 
reasoning process, allowing us to track how confidence evolves across the 
full trajectory rather than relying solely on the final output. Notably, this 
segmentation requires no additional generation: the steps $\{s_1, \ldots, s_T\}$ 
are extracted directly from the already-generated chain-of-thought, and each 
prefix $c_i$ is constructed by simple truncation.

\paragraph{Soft Probability Distribution.} 
To probe the model's implicit preference for each candidate answer at step $i$, 
we append a short answer-elicitation prompt to $c_i$ 
and perform a single forward pass to obtain next-token logits (see Appendix~\ref{app:prompt}). For each candidate 
answer token, the probability is computed as
\begin{equation}
q_i(a_j) = \frac{\exp(z_i^{(j)})}{\sum_{k=1}^{M} \exp(z_i^{(k)})}, 
\quad j = 1, \ldots, M,
\label{eq:qi}
\end{equation}
where $z_i^{(j)}$ is the logit corresponding to $a_j$ at step $i$. Our method 
requires only that each candidate answer can be disambiguated by its first token: 
in standard multiple-choice settings this is naturally satisfied by option labels 
(e.g., A, B, C, D), but the approach generalizes to any closed-set task where 
candidate answers have distinct first tokens, without requiring answers to be 
single tokens. We denote $q_i^* = q_i(\hat{a})$ as the probability assigned to the final predicted answer at step $i$.

The sequence $\{q_1^*, q_2^*, \ldots, q_T^*\}$ constitutes the \textbf{reasoning 
trajectory} in probability space. This training-free re-scoring of the already-generated 
chain-of-thought requires only $T$ lightweight forward passes, one per reasoning 
step, without additional text generation.

\subsection{STEP: Soft Trajectory Estimation of Prediction Correctness}

We define a family of step-level features that summarize different aspects of the 
trajectory's dynamics, each building on the previous by incorporating additional 
structure. All features aggregate information from the per-step distributions $q_i$ across steps and are normalized by the number of steps $T$, making them comparable across chains of different lengths.

\paragraph{Soft Consistency.} The simplest summary of the trajectory is the mean 
per-step confidence, which averages $q_i^*$ uniformly across all $T$ reasoning 
steps:
\[
\text{Con}_{\text{soft}} = \frac{1}{T} \sum_{i=1}^{T} q_i^*.
\]
This feature captures the overall level of commitment to the predicted answer 
throughout the reasoning process. Unlike hard consistency, which only records 
whether the predicted answer is the argmax at each step, $\text{Con}_{\text{soft}}$ 
preserves the full magnitude of $q_i^*$, providing a strictly richer signal.

\paragraph{Margin-Based Consistency.} Mean confidence alone does not capture how decisively the model distinguishes between competing candidate answers. To capture this decisiveness directly, we measure the gap between the top two probabilities at each step:
\[
\text{Con}_{\text{margin}} = \frac{1}{T} \sum_{i=1}^{T} 
\sigma\left(\alpha \left(q_i^{(1)} - q_i^{(2)}\right)\right),
\]
where $q_i^{(1)} \ge q_i^{(2)}$ are the two highest probabilities at step $i$, 
$\sigma$ is the sigmoid function, and $\alpha>0$ controls strictness. A large 
margin indicates the model decisively favors one answer over all others, while a 
small margin signals genuine uncertainty between competing options. In the following experiment section, we adopt $\alpha=5.0$.
\paragraph{Temporally-Weighted Consistency.}  As reasoning progresses, the model gradually eliminates competing 
options and its answer distribution concentrates toward the final prediction, 
making later steps more discriminative between correct and incorrect trajectories. 
We capture this by introducing an exponential weighting scheme that assigns higher 
weights to later steps:
\[
\text{Con}_{\text{temp}} = \frac{\sum_{i=1}^{T} w_i q_i^*}{\sum_{i=1}^{T} w_i}, 
\quad w_i = e^{\beta \cdot \frac{i}{T}}, \quad \beta>0,
\]
where $\beta$ controls the degree to which later steps are emphasized. As 
$\beta \to 0$, $\text{Con}_{\text{temp}}$ recovers $\text{Con}_{\text{soft}}$, 
making $\beta$ a natural generalization parameter.
\subsection{Relationship to Hard Indicators}

The soft features defined above are strictly more informative than the hard binary 
indicators used in prior trajectory-based methods, as they preserve the full 
continuous confidence signal at each step rather than reducing it to a binary 
correct/incorrect label.

The hard consistency score
\[
\text{Con}_{\text{hard}} = \frac{1}{T} \sum_{i=1}^{T} 
\mathbbm{1}\left[\hat{a} = \arg\max_{a_j} q_i(a_j)\right]
\]
retains only rank information and ignores the magnitude of the model's commitment 
at each step. Its soft counterpart, $\text{Con}_{\text{soft}}$, preserves the full 
probability value $q_i^*$: two steps with the same argmax but vastly different 
confidence levels are treated identically by $\text{Con}_{\text{hard}}$, yet carry 
very different information about reasoning reliability. $\text{Con}_{\text{margin}}$ 
and $\text{Con}_{\text{temp}}$ further enrich this signal by capturing decisiveness 
between competing options and temporal structure respectively, aspects that binary 
discretization fundamentally cannot encode.

Moreover, the closed-set structure of the task is what makes this single-trajectory
probing feasible in the first place: because $\mathcal{A}$ is predefined, we can
directly extract $q_i$ over a fixed candidate token set without resampling, whereas
prior work \citep{zhang2025covo} requires $N$ sampled trajectories to
construct a cross-trajectory distance matrix before any binary indicator can be 
computed. Our framework thus corresponds to a closed-set, single-trajectory 
simplification of the original CoVo consistency; the full derivation of this 
relationship is given in Appendix~\ref{app:covo}. 
\section{Experiments}
\begin{table*}[t]
\centering
\footnotesize
\renewcommand{\arraystretch}{1.05}
\setlength{\tabcolsep}{2pt}
\resizebox{\textwidth}{!}{%
\begin{tabular}{l ccc ccc ccc ccc}
\toprule
& \multicolumn{3}{c}{\textbf{MMLU}}
& \multicolumn{3}{c}{\textbf{CommonsenseQA}}
& \multicolumn{3}{c}{\textbf{Belebele}}
& \multicolumn{3}{c}{\textbf{ARC-Challenge}}\\
\cmidrule(lr){2-4}\cmidrule(lr){5-7}\cmidrule(lr){8-10}\cmidrule(lr){11-13}
\textbf{Method}
  & \scriptsize AUROC & \scriptsize AUPR-C & \scriptsize AUPR-E
  & \scriptsize AUROC & \scriptsize AUPR-C & \scriptsize AUPR-E
  & \scriptsize AUROC & \scriptsize AUPR-C & \scriptsize AUPR-E
  & \scriptsize AUROC & \scriptsize AUPR-C & \scriptsize AUPR-E\\
\midrule
\multicolumn{13}{l}{\textsc{Qwen2.5-7B-Instruct}} \\[2pt]
\multicolumn{13}{l}{\textit{Final-output statistics}} \\
\quad Entropy         & 54.00 & 72.44 & 32.43 & 56.96 & 83.95 & 23.42 & 55.39 & 91.04 & 12.04 & 58.81 & 90.05 & 16.64\\
\quad Perplexity      & 53.39 & 72.19 & 31.91 & 56.09 & 83.41 & 22.74 & 54.94 & 90.06 & 11.80 & 58.44 & 90.20 & 15.84\\
\quad Max Prob.       & 53.33 & 72.11 & 31.93 & 56.07 & 83.35 & 22.92 & 54.69 & 90.88 & 11.76 & 58.62 & 90.27 & 16.03\\[2pt]
\multicolumn{13}{l}{\textit{Latent-trajectory}} \\
\quad CoE-R           & 64.37 & 80.58 & 41.43 & 55.50 & 82.37 & 23.89 & 55.27 & 90.87 & 14.76 & 57.32 & 89.05 & 16.20\\
\quad CoE-C           & 64.87 & 80.52 & 41.73 & 56.07 & 82.00 & 26.05 & 55.83 & 90.81 & 14.29 & 62.17 & 90.59 & 20.74\\
\quad CoT-Kinetics    & 59.99 & 75.79 & 37.44 & 53.49 & 81.59 & 22.68 & 54.71 & 90.33 & 12.21 & 60.14 & 90.34 & 17.22\\[2pt]
\multicolumn{13}{l}{\textit{Hard token-trajectory}} \\
\quad CoVo-Con        & 70.58 & 86.49 & 53.85 & 73.01 & 92.25 & 45.43 & 78.72 & 97.23 & \textbf{49.76} & 64.84 & 94.91 & 34.55\\
\quad CoVo-Vol        & 70.45 & 86.44 & 54.00 & 73.00 & 92.23 & \textbf{46.83} & 77.95 & 97.15 & 44.31 & 64.92 & 94.92 & \underline{36.23}\\
\quad CoVo-Linear     & 70.73 & 86.52 & 54.32 & 73.27 & \underline{92.29} & 46.10 & 78.85 & 97.21 & \underline{48.45} & 64.98 & 94.93 & 35.66\\[2pt]
\multicolumn{13}{l}{\textit{STEP (ours)}} \\
\quad $\textsc{Con}_\text{soft}$   & \underline{75.23} & \underline{86.91} & \underline{55.68} & \underline{76.06} & 92.25 & 46.44 & \underline{83.60} & \underline{97.40} & 46.94 & \underline{77.97} & \textbf{95.56} & 35.48\\
\quad $\textsc{Con}_\text{margin}$ & 74.22 & 86.48 & 54.90 & 75.56 & 91.98 & 45.95 & 82.87 & 97.30 & 46.81 & 77.66 & \underline{95.53} & 35.46\\
\quad $\textsc{Con}_\text{temp}$   & \textbf{75.92} & \textbf{87.24} & \textbf{56.82} & \textbf{76.86} & \textbf{92.50} & \underline{46.76} & \textbf{84.16} & \textbf{97.45} & 46.54 & \textbf{78.38} & 95.49 & \textbf{37.01}\\
\midrule
\multicolumn{13}{l}{\textsc{Meta-Llama-3-8B-Instruct}} \\[2pt]
\multicolumn{13}{l}{\textit{Final-output statistics}} \\
\quad Entropy         & 49.84 & 59.63 & 40.16 & 48.90 & 72.53 & 25.82 & 53.61 & 90.93 & 11.23 & 48.51 & 78.53 & 20.40\\
\quad Perplexity      & 49.49 & 59.40 & 39.74 & 49.12 & 72.78 & 26.08 & 53.65 & 90.78 & 11.18 & 48.67 & 78.35 & 20.25\\
\quad Max Prob.       & 49.43 & 59.34 & 39.70 & 48.94 & 72.68 & 26.07 & 53.76 & 90.74 & 11.16 & 48.28 & 78.27 & 19.99\\[2pt]
\multicolumn{13}{l}{\textit{Latent-trajectory}} \\
\quad CoE-R           & 59.21 & 68.68 & 48.05 & 54.21 & 75.15 & 29.38 & 59.69 & 91.81 & 14.32 & 56.52 & 82.91 & 26.43\\
\quad CoE-C           & 53.56 & 64.01 & 42.17 & 51.19 & 74.18 & 26.69 & 52.93 & 89.93 & 11.92 & 56.05 & 82.76 & 25.38\\
\quad CoT-Kinetics    & 57.35 & 67.32 & 44.92 & 52.85 & 74.86 & 28.26 & 54.25 & 89.76 & 12.52 & 57.76 & 83.73 & 26.24\\[2pt]
\multicolumn{13}{l}{\textit{Hard token-trajectory}} \\
\quad CoVo-Con        & 77.54 & \underline{86.34} & 74.12 & 77.29 & \textbf{91.55} & 66.29 & 87.33 & 98.39 & \textbf{68.95} & 84.52 & 95.65 & 72.95\\
\quad CoVo-Vol        & 76.91 & 86.12 & 73.74 & 75.86 & 91.11 & 64.91 & 86.86 & 98.34 & 67.20 & 84.29 & 95.61 & 73.26\\
\quad CoVo-Linear     & 77.52 & 86.32 & 74.11 & 77.27 & \underline{91.54} & 66.00 & 87.29 & 98.38 & \underline{68.80} & 84.60 & 95.67 & 73.32\\[2pt]
\multicolumn{13}{l}{\textit{STEP (ours)}} \\
\quad $\textsc{Con}_\text{soft}$   & \underline{82.11} & 86.28 & \underline{74.73} & \underline{81.24} & 90.93 & 66.82 & \underline{91.78}& \underline{98.62}& 67.76 & \underline{88.83} & \underline{96.15} & \underline{74.94}\\
\quad $\textsc{Con}_\text{margin}$ & 81.75 & 86.07 & 73.96 & 81.03 & 90.93 & \underline{67.01} & 91.56 & 98.60 & 64.04 & 88.51 & 96.06 & 73.47\\
\quad $\textsc{Con}_\text{temp}$   & \textbf{82.29} & \textbf{86.47} & \textbf{74.84} & \textbf{81.70} & 91.20 & \textbf{67.38} & \textbf{91.89} & \textbf{98.64} & 67.28 & \textbf{89.24} & \textbf{96.24} & \textbf{76.09}\\
\bottomrule
\end{tabular}%
}
\caption{AUROC, AUPR-C, and AUPR-E on LLM experiments across four benchmarks
and two models. \textbf{Bold}: best result per column within each model block. 
\underline{Underline}: second best result per column within each model block.}
\label{tab:main_llm}
\end{table*}

\begin{figure*}[t]
    \centering
    \includegraphics[width=\linewidth]{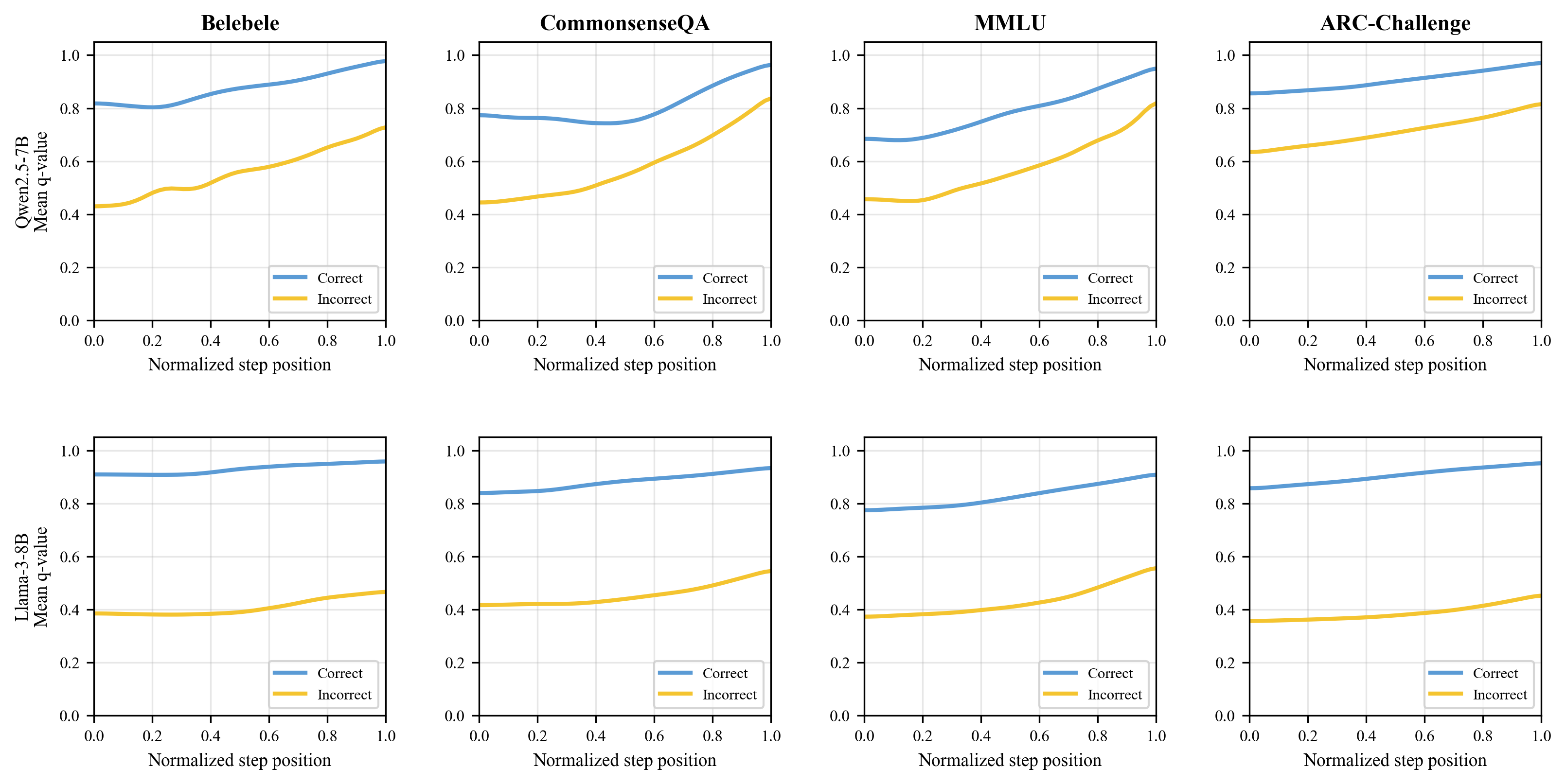}
    \caption{Mean $q^*$ trajectory across normalized reasoning steps for correct 
    (blue) and incorrect (orange) predictions, across four datasets and two models. Correct trajectories consistently maintain higher confidence and 
    converge toward near-certainty by the final step, while incorrect trajectories 
    plateau at a lower level. The widening gap toward later steps motivates the 
    temporal weighting in $\text{Con}_{\text{temp}}$.}
    \label{fig:trajectory}
\end{figure*}

\subsection{Experimental Setup}
\label{sec:setup}
\paragraph{Datasets.}
We evaluate on four datasets spanning multiple reasoning domains: 
\textbf{MMLU}~\citep{hendrycks2021mmlu}, a massive multitask benchmark covering 57 academic 
subjects; \textbf{CommonsenseQA}~\citep{talmor2019commonsenseqa}, a commonsense reasoning 
benchmark with commonsense-related multiple-choice questions; \textbf{Belebele}~\citep{bandarkar2023belebele}, 
a comprehensive reading comprehension benchmark; and \textbf{ARC-Challenge}~\citep{clark2018arc}, 
a set of grade-school level science questions requiring complex reasoning.
Dataset statistics are provided in Appendix~\ref{app:datasets}.

\paragraph{Models.}
We select \textbf{Qwen2.5-7B-Instruct}~\citep{yang2024qwen25} and 
\textbf{Meta-Llama-3-8B-Instruct}~\citep{grattafiori2024llama3}, two representative open-source 
instruction-tuned models at the 7--8B parameter scale. We evaluate on models of this 
scale as they offer a practical 
balance between reasoning capability and computational efficiency. All models
generate chain-of-thought reasoning by prompting them to explicitly articulate
intermediate reasoning steps before producing a final answer, following the prompt
template described in Appendix~\ref{app:cot_prompt}. For each query we generate
\emph{one} fixed chain-of-thought rollout per model and reuse it for every
method compared, so differences across methods reflect only the scoring
procedure, not generation variance.

\paragraph{Task Accuracy.}
Table~\ref{tab:acc_cost} reports task accuracy for the same generations used
throughout. Accuracy ranges
61.13--89.89\%, confirming the task is meaningfully imbalanced
(Section~\ref{sec:setup}) rather than near-ceiling or near-chance. At most
0.85\% of generations lack a parsable answer and are excluded.

\begin{table}[t]
\centering
\footnotesize
\setlength{\tabcolsep}{4pt}
\begin{tabular}{lcc}
\toprule
\textbf{Dataset} & Qwen2.5-7B & Llama-3-8B \\
 & Acc.\,(\%) & Acc.\,(\%) \\
\midrule
MMLU          & 70.60 & 61.13 \\
CommonsenseQA & 80.67 & 74.42 \\
Belebele      & 89.89 & 89.42 \\
ARC-Challenge & 87.63 & 79.20 \\
\bottomrule
\end{tabular}
\caption{Task accuracy per benchmark and model.}
\label{tab:acc_cost}
\end{table}

\paragraph{Computational Cost.}
STEP's only overhead versus the free final-output-statistics baselines is
$T$ lightweight forward passes per query, no new tokens generated
(Appendix~\ref{app:prompt}); $\bar{T}$ remains small across all datasets
(typically in the single digits). This matches the hard token-trajectory baselines (CoVo-Con/Vol/Linear), which
use the same $q_i$, so the soft-vs-hard gain (Section~\ref{sec:main_results})
is free. STEP's real advantage is over methods needing $N$ \emph{sampled} full
CoT generations, e.g.\ original CoVo~\citep{zhang2025covo}
(Appendix~\ref{app:covo}) and semantic uncertainty
methods~\citep{kuhn2023semantic, farquhar2024detecting}: $O(N\bar{T})$
generated tokens vs.\ STEP's $O(T)$ forward passes with none. More information is in Appendix~\ref{app:compute}.

\paragraph{Baselines.}
We compare against three categories of baselines.
(1)~\textit{Final-output statistics}: \textbf{Entropy}, token-level 
perplexity (\textbf{Perplexity}), and maximum softmax probability (\textbf{MaxProb}), 
which aggregate token-level statistics over the entire generated sequence and 
discard the reasoning structure entirely.
(2)~\textit{Latent-trajectory methods}: \textbf{CoE-R} and 
\textbf{CoE-C}~\citep{wang2024coe}, which derive correctness scores from the geometric 
trajectory of hidden states across layers, and \textbf{CoT-Kinetics}~\citep{bi2025cotkinetics}, 
which models the dynamics of hidden state transitions across reasoning steps.
(3)~\textit{Hard token-trajectory methods}: \textbf{CoVo-Con}, \textbf{CoVo-Vol}, 
and \textbf{CoVo-Linear}, which are adaptations of the consistency and volatility 
concepts introduced by \citet{zhang2025covo} to our single-trajectory setting.
These variants derive confidence from per-step binary correctness indicators 
rather than soft probabilities, discarding fine-grained confidence information.

\paragraph{Evaluation Metrics.}
To better evaluate the quality of different scoring methods, we report AUROC~\citep{fawcett2006roc}, AUPR-C and AUPR-E~\citep{boyd2013area} as evaluation metrics, 
following established protocols for misclassification 
detection~\citep{hendrycks2017baseline}.
AUROC (Area Under the Receiver Operating Characteristic curve) measures the overall 
discriminative ability between correct and incorrect predictions.
AUPR-C (Area Under the Precision-Recall curve for the \textit{Correct} class) and 
AUPR-E (Area Under the Precision-Recall curve for the \textit{Error} class) treat 
correct and incorrect predictions as the positive class, respectively.
Since LLM accuracy on our benchmarks ranges from 60\% to 90\%, the class distribution 
is heavily skewed toward correct predictions, making AUROC alone 
insufficient~\citep{corbiere2019failure}.
AUPR-C reflects the ability to confirm high-confidence correct predictions, while 
AUPR-E captures the more practically important ability to identify errors among the 
minority class.
Formal definitions are provided in Appendix~\ref{app:metrics}.

\subsection{Main Results}
\label{sec:main_results}

Table~\ref{tab:main_llm} presents AUROC, AUPR-C, and AUPR-E of all methods on 
four benchmarks and two models.

\paragraph{STEP consistently outperforms all baselines.}
Across all datasets and both models, soft trajectory features achieve 
the best overall performance, surpassing output-statistics-based, latent-trajectory, and 
hard token-trajectory baselines in the vast majority of settings. Although 
individual metrics occasionally favor a specific baseline on a particular 
dataset, the improvements are consistent across models and tasks, validating 
that preserving continuous probability signals across reasoning steps provides 
richer information for correctness estimation than either final-output 
statistics or hard trajectory indicators.

\paragraph{Weighting favoring later steps further boosts performance.}
Among the three variants of STEP, \textit{Temporally-Weighted Consistency} $\text{Con}_{\text{temp}}$ achieves the
best evaluation metrics in nearly all settings, outperforming the simpler
\textit{Soft Consistency} $\text{Con}_{\text{soft}}$ and \textit{Margin Consistency} $\text{Con}_{\text{margin}}$; we recommend it as the single deployed estimator, reporting the other two as ablations rather than alternatives to combine at inference time. This suggests that not all reasoning steps
contribute equally to correctness estimation: later steps, where the model's answer
distribution has begun to stabilize, carry more predictive signal.
$\text{Con}_{\text{soft}}$ and $\text{Con}_{\text{margin}}$ still substantially
outperform hard token-trajectory methods despite treating all steps uniformly,
confirming the gain from soft signals is orthogonal to the aggregation strategy.
\subsection{Trajectory Analysis}

Figure~\ref{fig:trajectory} visualizes the mean $q^*$ trajectory across reasoning 
steps for correct and incorrect predictions, averaged over all samples in each 
dataset. Correct trajectories maintain substantially higher $q^*$ values throughout 
the reasoning process, confirming that the model's commitment to its final answer 
is a reliable signal of correctness. This pattern is consistent across all four datasets and 
both model families, suggesting it reflects a general property of LLM reasoning 
rather than a dataset- or model-specific artifact. The base variant of STEP, 
\textit{Soft Consistency}, which simply averages $q^*$ throughout the 
chain-of-thought, is therefore shown to be effective.

\subsection{Ablation Study}

\begin{figure}[t]
    \centering
    \includegraphics[width=0.7\linewidth]{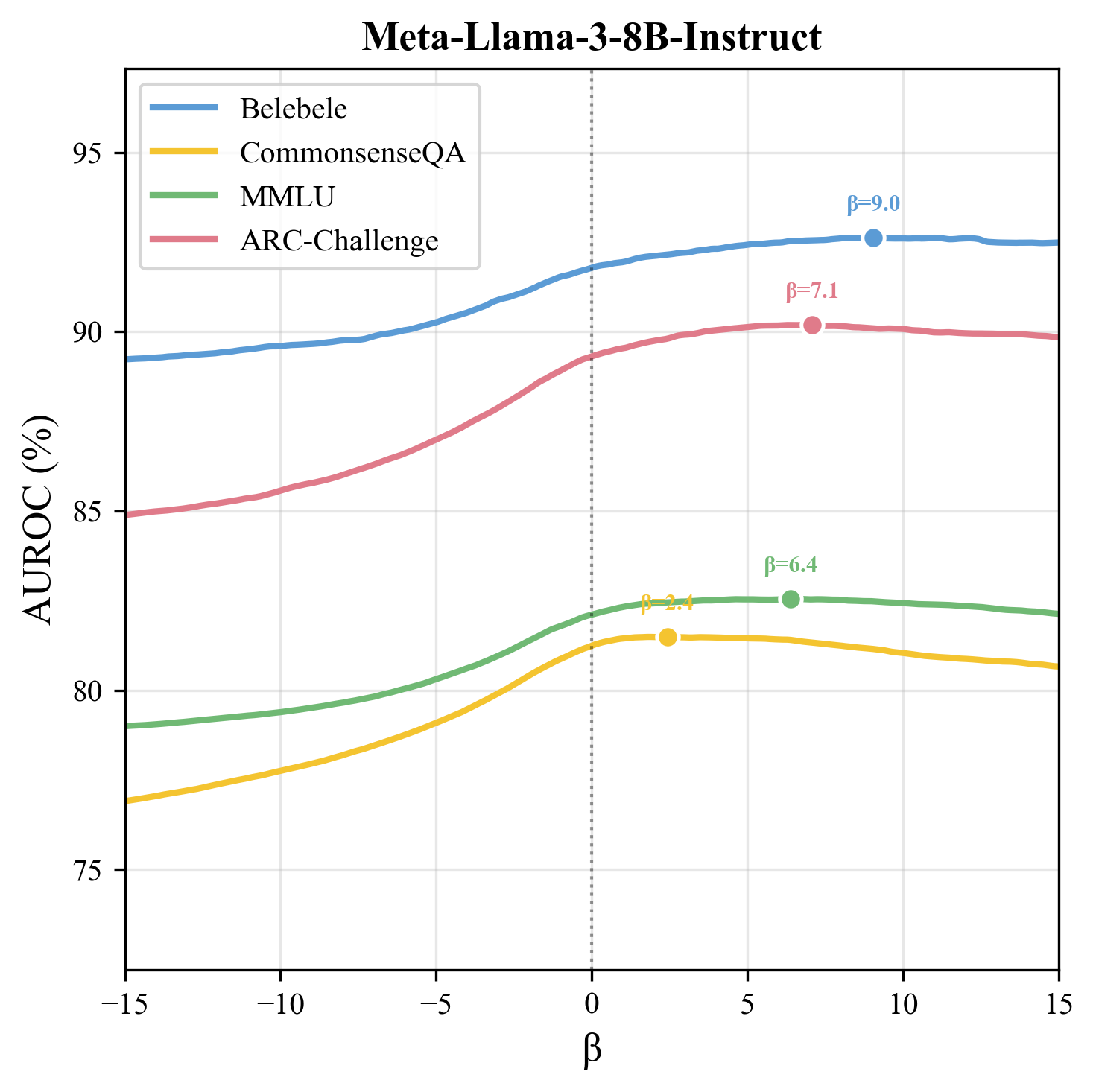}
    \caption{AUROC as a function of the temporal weighting parameter $\beta$ 
across datasets on Meta-Llama-3-8B-Instruct. $\beta > 0$ up-weights later 
reasoning steps; $\beta = 0$ recovers $\text{Con}_{\text{soft}}$.}
    \label{fig:beta}
\end{figure}

Figure~\ref{fig:beta} illustrates how the temporal weighting parameter $\beta$ 
affects AUROC across datasets on model Meta-Llama-3-8B-Instruct. Recall that $\beta > 0$ assigns 
higher weights to later reasoning steps, $\beta < 0$ to earlier steps, and 
$\beta = 0$ recovers $\text{Con}_{\text{soft}}$.

Across all datasets, performance increases monotonically as $\beta$ grows from 
negative to positive values. We highlight three observations:

\textbf{Later steps are more informative.} Performance improves consistently as 
$\beta$ increases from zero, confirming that reasoning steps closer to the final 
answer carry stronger correctness signal. This aligns with the intuition that early steps often consist of problem restatement or initial exploration, where the model has not yet committed to any reasoning direction. As reasoning progresses, the model gradually eliminates competing options and its answer distribution concentrates toward the final prediction, producing the largest divergence between correct and incorrect trajectories. Temporal weighting respects this process by giving the model room to explore early on while placing greater emphasis on the steps where meaningful commitment emerges.

\textbf{Performance saturates at large $\beta$.} Gains of AUROC plateau around 
$\beta \approx 5$ to $10$ and remain stable thereafter. We attribute this to a 
ceiling effect at the very final steps: regardless of whether the prediction is 
correct or incorrect, the model tends to assign near-certain probability to its 
predicted answer as it self-consistently converges toward its own conclusion, 
reducing the discriminative value of the last steps.

\textbf{Negative $\beta$ consistently hurts performance.} To further verify that 
later steps are more informative than earlier ones, we also explore negative values 
of $\beta$, which up-weight early steps instead. Performance degrades consistently 
across all datasets, confirming our hypothesis: early steps, where the model has 
not yet begun to commit to any candidate and the answer distribution remains noisy, 
carry weaker signal than the later steps where meaningful reasoning commitment 
emerges.

\section{Conclusion}
We presented STEP, a training-free framework for correctness estimation in 
closed-set reasoning tasks. By performing lightweight forward passes on an 
already-generated chain-of-thought, STEP derives soft trajectory features 
from per-step probability distributions over the candidate answer set. 
Across multiple benchmarks and two models, all three instantiations of STEP consistently outperform existing baselines in AUROC while achieving competitive or improved AUPR performance. Among them, \textit{Temporally-Weighted 
Consistency} achieves the strongest performance, confirming that later
reasoning steps carry disproportionately more predictive signal. These
results validate the core premise of STEP: preserving continuous probability 
signals across reasoning steps provides richer information than hard binary 
indicators, at no additional generation cost. Future work may explore 
extending soft trajectory features to open-ended settings by constructing 
candidate sets dynamically from the model's top-$k$ predictions at each step.

\section*{Limitations}

\paragraph{Scope of applicability.}
STEP is designed for closed-set tasks with a predefined candidate answer set, 
and is not directly applicable to open-ended generation. It further requires 
that candidate answers be disambiguable by their first token; answers sharing 
a common prefix would require additional aggregation strategies.

\paragraph{Dependence on reasoning structure.}
STEP relies on structured chain-of-thought reasoning to construct per-step 
prefixes. Models producing unstructured or highly variable reasoning, as well 
as formal symbolic tasks where correctness depends on exact computation rather 
than progressive semantic narrowing, may not benefit consistently.

\paragraph{Access requirements.}
STEP requires access to next-token logits, available for open-source models 
but not for black-box APIs.
\bibliography{references}

\appendix
\section{Answer-Elicitation Prompt}
\label{app:prompt}

To compute the per-step answer distribution $q_i$ (Equation~\ref{eq:qi}), 
we append the following prompt to the truncated context $c_i$ before 
performing the forward pass:

\begin{verbatim}
Final Answer:
\end{verbatim}

This prompt is designed to elicit the model's implicit preference over 
candidate answers at the current reasoning state, without generating any 
additional reasoning content. By appending it immediately 
after the truncated chain-of-thought, we place the model in a position where 
its next-token distribution directly reflects its answer preference given 
everything it has reasoned so far. The logits over the candidate answer tokens 
 are extracted at this position and normalized via 
softmax over the candidate set to obtain $q_i$, as described in 
Equation~\ref{eq:qi}. No further tokens are generated beyond this single 
forward pass.
\section{Chain-of-Thought Generation Prompt}
\label{app:cot_prompt}

All models are prompted with the following template to elicit structured
chain-of-thought reasoning. The placeholder \texttt{\{input\_data\}} is replaced
with the dataset-specific question and answer choices.
The candidate set is \{A, B, C, D, E\} for CommonsenseQA and
\{A, B, C, D\} for all other datasets.

\vspace{0.5em}
\noindent\fbox{%
\begin{minipage}{0.95\linewidth}
\ttfamily\small
Answer the following multiple choice question.

\medskip
Explain your reasoning in a few statements.\\
Each statement should represent \textbf{one small logical improvement} over the previous statements.\\
Write each statement on its own line.\\
Do not number or label the statements.

\medskip
The final line must be:\\
Answer: X

\medskip
Where X MUST BE one of A, B, C, D [, or E].\\
Do not write anything after the answer.

\medskip
\{input\_data\}
\end{minipage}%
}

\vspace{0.5em}
\noindent
The prompt deliberately avoids numbered lists or structured headers to produce
free-form reasoning steps that can be cleanly segmented by newline splitting.
Each line of the response (excluding the final \texttt{Answer:} line) constitutes
one reasoning step $s_k$, yielding a trajectory of length $T$ for confidence scoring.
\section{Evaluation Metrics}
\label{app:metrics}

\paragraph{AUROC.}
The Area Under the Receiver Operating Characteristic Curve measures the 
probability that a randomly chosen correct prediction receives a higher 
correctness score than a randomly chosen incorrect prediction:
\begin{equation}
  \text{AUROC} = \mathbb{P}\!\left(\phi(x^+) > \phi(x^-)\right)
\end{equation}
where $x^+$ and $x^-$ are randomly drawn correct and incorrect samples 
respectively, and $\phi(\cdot)$ denotes the correctness score assigned by 
the method. A value of 1.0 indicates perfect separation; 0.5 corresponds 
to random performance.

\paragraph{AUPR-C.}
The Area Under the Precision-Recall Curve with correct predictions as the 
positive class summarizes the trade-off between precision and recall across 
all decision thresholds:
\begin{equation}
  \text{AUPR-C} = \int_0^1 P_C(r)\, dr
\end{equation}
where $P_C(r)$ is the precision at recall level $r$ when treating correct 
predictions as positives. AUPR-C reflects the ability to confirm 
high-confidence correct predictions, and serves as a natural complement to 
AUROC under class imbalance.

\paragraph{AUPR-E.}
The Area Under the Precision-Recall Curve with incorrect predictions as the 
positive class measures the ability to detect errors among the minority class:
\begin{equation}
  \text{AUPR-E} = \int_0^1 P_E(r)\, dr
\end{equation}
where $P_E(r)$ is the precision at recall level $r$ when treating incorrect 
predictions as positives. Since LLM accuracy on our benchmarks ranges from 
60\% to 90\%, incorrect predictions constitute a small minority, making 
AUPR-E particularly informative for evaluating error detection capability 
in practical deployment scenarios.

\section{Dataset Statistics}
\label{app:datasets}

We evaluate on four multiple-choice question answering datasets spanning diverse
reasoning domains. MMLU~\citep{hendrycks2021mmlu} covers 57 academic subjects ranging
from STEM to humanities. CommonsenseQA~\citep{talmor2019commonsenseqa} requires
everyday commonsense knowledge to disambiguate between plausible answer choices.
Belebele~\citep{bandarkar2023belebele} tests reading comprehension across typologically
diverse languages; we use the English subset. ARC-Challenge~\citep{clark2018arc}
consists of science questions specifically selected to be resistant to simple
retrieval-based and word co-occurrence methods.
Table~\ref{tab:dataset_stats} summarizes the key statistics.

\begin{table}[h]
\centering
\small
\begin{tabular}{lcc}
\toprule
\textbf{Dataset} & \textbf{\#Samples} & \textbf{\#Choices} \\
\midrule
MMLU          & 1{,}531 & 4 \\
CommonsenseQA & 2{,}000 & 5 \\
Belebele      &    900  & 4 \\
ARC-Challenge & 1{,}172 & 4 \\
\bottomrule
\end{tabular}
\caption{Dataset statistics used in our evaluation.}
\label{tab:dataset_stats}
\end{table}

\section{Relationship to CoVo Consistency}
\label{app:covo}

The original CoVo consistency \citep{zhang2025covo} is defined over $N$ sampled 
trajectories with $K$ distinct answers via a distance matrix 
$\mathbf{D} \in \mathbb{R}^{T \times K}$, where
\begin{align*}
\mathbf{D}[i,k] &= d(s_i, y_k) \\
&= -\frac{1}{|y_k|}\sum_{j=1}^{|y_k|}
\log \pi_\theta(y_k[j] \mid s_i, y_k[:j])
\end{align*}
is the negative log-likelihood of answer $y_k$ at reasoning state $s_i$. 
Consistency is then defined as the fraction of intermediate states whose own 
final answer is closer than any alternative:
\begin{equation}
\text{Con}_{\text{CoVo}}(\tau)
=
\frac{1}{T}\sum_{i=0}^{T-1}
\mathbbm{1}\!\left[
\mathbf{D}[i,0]
=
\min_{0 \le k < K}\mathbf{D}[i,k]
\right]
\end{equation}
This formulation requires $N$ sampled trajectories to populate $K$ distinct 
candidate answers, since the alternative answers $\{y_1, \ldots, y_{K-1}\}$ 
are obtained from other sampled trajectories rather than from a predefined set.

In the closed-set setting, $\mathcal{A} = \{a_1, \ldots, a_M\}$ is predefined 
and fixed, which admits two simplifications. First, the candidate set is known 
without resampling, so the distance matrix can be replaced by the per-step 
softmax distribution $q_i$ defined in Equation~\ref{eq:qi}, where 
$q_i(a_j) \propto \exp(z_i^{(j)})$ directly reflects the model's relative 
preference over all candidates at step $i$. Second, since $q_i$ is already 
normalized over $\mathcal{A}$, the distance-based argmax comparison reduces to:
\begin{align*}
&\mathbf{D}[i,0] = \min_{0 \le k < K} \mathbf{D}[i,k] \\
&\iff \hat{a} = \arg\max_{a_j} q_i(a_j)
\end{align*}
Applying these substitutions to $\text{Con}_{\text{CoVo}}$ and dropping the 
group-level summation over $G$ trajectories (as only a single trajectory is 
available) recovers exactly $\text{Con}_{\text{hard}}$ as defined in 
Section~\ref{sec:method}:
\[
\text{Con}_{\text{hard}} = \frac{1}{T} \sum_{i=1}^{T} 
\mathbbm{1}\!\left[\hat{a} = \arg\max_{a_j} q_i(a_j)\right]
\]
where we reindex intermediate states from $i=1$ to $T$ to match the notation 
used throughout this paper; the original formulation uses $i=0$ to $T-1$.
$\text{Con}_{\text{hard}}$ thus inherits the binary structure of 
$\text{Con}_{\text{CoVo}}$ while requiring only a single forward pass per 
reasoning step. The soft features introduced in Section~\ref{sec:method} 
replace this binary indicator with the continuous value $q_i^*$, preserving 
the full confidence signal that the hard indicator discards.

\section{Runtime and Hardware}
\label{app:compute}
All experiments were run on a single NVIDIA A100 GPU, with no model training
involved. Total inference time per dataset, including the initial
chain-of-thought generation and all downstream scoring (final-output
statistics, latent-trajectory, hard token-trajectory, and STEP), ranges from
20 minutes to approximately 3--4 hours depending on dataset size and model.

Each of STEP's additional forward passes (Section~\ref{sec:setup}) processes a fixed truncated prefix and produces no additional output tokens. Unlike sampling-based methods, STEP therefore avoids autoregressive generation of additional full reasoning trajectories.

\end{document}